\documentclass[
twocolumn,
]{ceurart}

\usepackage{listings}
\usepackage{adjustbox}
\graphicspath{ {./figures/} }

\begin{document}

\copyrightyear{2023}
\copyrightclause{Copyright for this paper by its authors.
  Use permitted under Creative Commons License Attribution 4.0
  International (CC BY 4.0).}

\conference{AutomationXP23: Intervening, Teaming, Delegating Creating Engaging Automation Experiences, April 23rd, Hamburg, Germany}

\title{ReBound: An Open-Source 3D Bounding Box Annotation Tool for Active Learning}


\author{Wesley Chen}
\fnmark[1]

\author{Andrew Edgley}
\fnmark[1]

\author{Raunak Hota}
\fnmark[1]

\author{Joshua Liu}
\fnmark[1]

\author{Ezra Schwartz}
\fnmark[1]

\author{Aminah Yizar}
\fnmark[1]

\author{Neehar Peri}
\fnmark[2]

\author{James Purtilo}
\fnmark[2]

\fntext[1]{These authors contributed equally.}
\fntext[2]{These authors advised equally.}
\cortext[1]{Corresponding author: \href{mailto:purtilo@umd.edu}{purtilo@umd.edu}}

\begin{abstract}
In recent years, supervised learning has become the dominant paradigm for training deep-learning based methods for 3D object detection. Lately, the academic community has studied 3D object detection in the context of autonomous vehicles (AVs) using publicly available datasets such as nuScenes and Argoverse 2.0. However, these datasets may have incomplete annotations, often only labeling a small subset of objects in a scene. Although commercial services exists for 3D bounding box annotation, these are often prohibitively expensive. To address these limitations, we propose ReBound, an open-source 3D visualization and dataset re-annotation tool that works across different datasets. In this paper, we detail the design of our tool and present survey results that highlight the usability of our software. Further, we show that ReBound is effective for exploratory data analysis and can facilitate active-learning. Our code and documentation is available on \href{https://github.com/ajedgley/ReBound}{GitHub}.

\end{abstract}

\begin{keywords}
Autonomous Driving, Active Learning, 3D Annotation Tools, Human Computer Interaction, Data Visualization
\end{keywords}

\maketitle
\section{Introduction}
3D object detection is a critical component of the autonomous vehicle (AV) perception stack ~\cite{geiger2012we, caesar2020nuscenes}. To facilitate research in 3D object detection, the AV industry has released large-scale 3D annotated multimodal datasets~\cite{caesar2020nuscenes, wilson2021argoverse, sun2020waymo}. These datasets include LiDAR sweeps and multi-camera RGB images from diverse driving logs, which captures detailed information about the surrounding environment. Crucially, objects of interest are annotated by drawing 3D bounding boxes and labeling them as part of a particular category. Contemporary 3D detectors \cite{yin2020center, peri2022towards, lang2019pointpillars, zhu2019class, peri2022futuredet} are trained using supervised learning, and are limited by the annotations provided with the dataset. For example, objects in the nuScenes dataset are inconsistently labeled beyond 50m, making it challenging to evaluate long-range detection. Further, the nuScenes dataset does not label street signs and traffic lights, which are critical for safe navigation. 

\begin{figure*}[t]
    \centering
    \includegraphics[width=\linewidth]{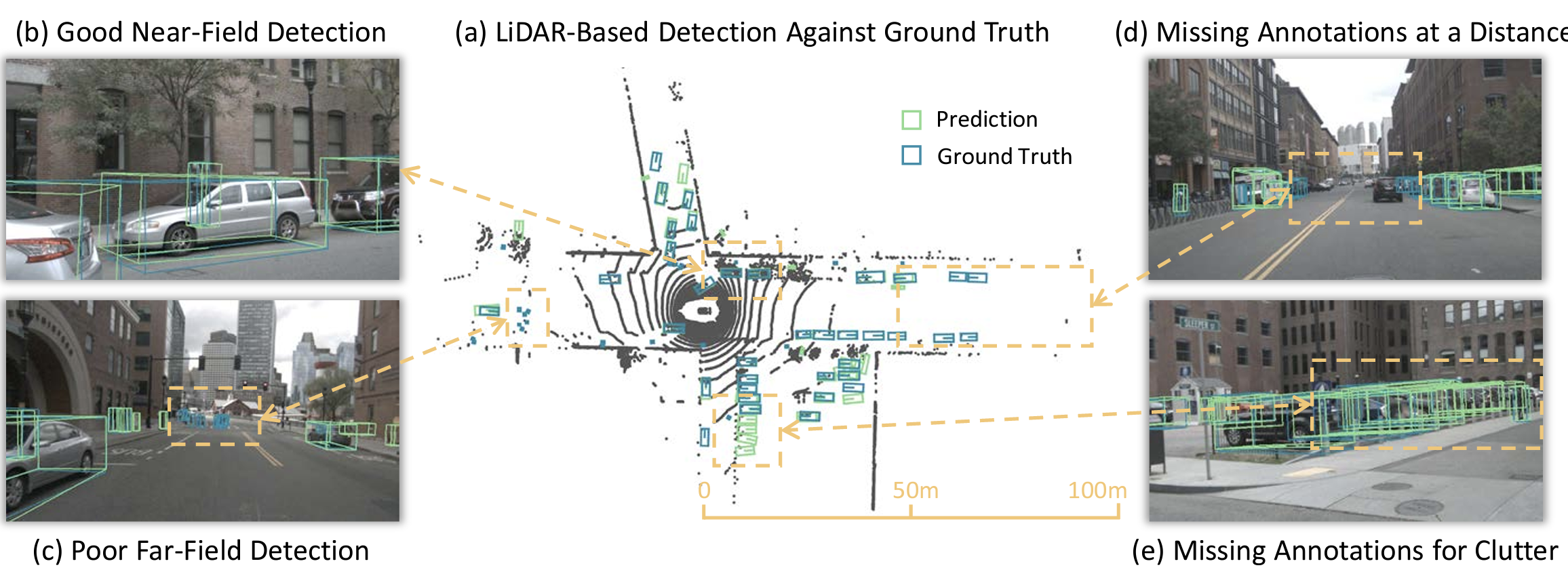}
    \caption{Long-range 3D detection is crucial for safe navigation in autonomous vehicles. Modern benchmarks have greatly improved 3D object detection. However, standard benchmarks like nuScenes only evaluate detections up to 50m, masking the poor far-field object detection performance of 3D detectors (b \& c). Existing benchmarks do not sufficiently annotate far-field objects (d \& e), partly due to fewer LiDAR returns. Our proposed annotation framework facilitates far-field annotation, and will enable researchers to better study this problem. We adopt this figure from \cite{gupta2023far3det}.}
    \label{fig:missing_anno}
\end{figure*}

Manually re-annotating datasets with new 3D bounding box annotations is particularly challenging. Annotating 3D bounding boxes using 2D RGB images is difficult because it is not possible to accurately estimate bounding-box depth. Similarly, annotating 3D bounding boxes using LiDAR point clouds is difficult because LiDAR returns are sparse, making it tough to identify individual objects at long-range and in cluttered scenes, as shown in Fig. \ref{fig:missing_anno}. Although commercial services can annotate 3D data at scale, it is often prohibitively expensive. As a result, several tools have been created for quick, efficient, and easy point cloud annotation \cite{sager21_labelcloud, arief20_sane, wang19_latte, zimmer19_bat}, with the intention of making it simpler for researchers to create their own datasets. However, these tools do not support a wide variety of data formats, multi-modal exploratory data analysis and active learning. 


\begin{figure}[t]
    \centering
    \includegraphics[scale=0.25]{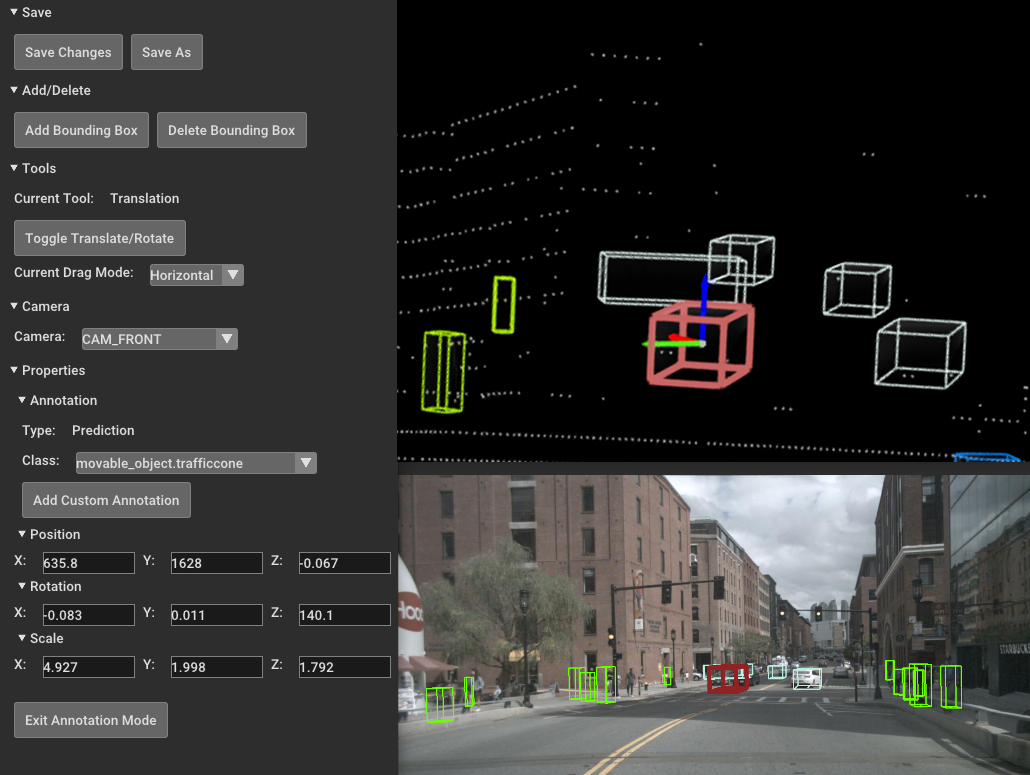}
    \caption{We present ReBound, an open-source 3D annotation tool that allows users to add, delete, and modify annotations from existing datasets or model predictions to support active learning.}
    \label{fig:rebound}
\end{figure}

In this paper, we introduce ReBound, an open-source annotation tool which aims to simplify the process of multi-modal 3D visualization and bounding box annotation. Our framework is designed to provide users with the ability to import and modify annotations from various datasets. To achieve this, we propose a generic data type that can be extended to accommodate different data formats. Additionally, ReBound enables active learning by allowing users to correct bounding box predictions and labels, create new custom labels, analyze model predictions, and export new annotations back to dataset-specific formats for model re-training. We measure the effectiveness of our tool through user studies, focusing on the ease of using the data conversion and annotation editing features. Our results show that the tool is both intuitive to use and useful for rapidly adding new annotations.

\section{Related Work}

In this section we present a brief overview of existing 3D annotation tools and active learning methods for object detection.

\textbf{3D Annotation Tools.}
 Modern 3D detectors require diverse, large-scale 3D annotations for supervised learning. A variety of tools have been created to address this challenge \cite{li20_autochallenges}, including labelCloud\cite{sager21_labelcloud}, SAnE\cite{arief20_sane}, LATTE\cite{wang19_latte}, and 3D BAT\cite{zimmer19_bat}, which all allow for efficient manual or semi-automatic dataset annotation.
However, unlike these existing tools, ReBound also allows users to manually update annotations from existing datasets for evolving use cases like detecting a new category or updating annotations for far-field objects. 
Our tool currently supports re-annotation for the nuScenes\cite{caesar20_nuscenes}, Waymo \cite{sun20_waymo}, and Argoverse 2.0\cite{wilson21_argo} datasets.

\textbf{Active Learning for 3D Object Detection.}
Although manually labeling tens of thousands of images or LiDAR sweeps can be prohibitively expensive, recent work in active learning suggests that we can bootstrap deep learning models by iteratively annotating and training on a targeted subset of informative examples to significantly improve model performance. Selecting the most relevant data to be annotated by human experts is a primary challenge for active learning. Recent works define a scoring function, using either an uncertainty-based \cite{haussmann20_activeobj, MIAOD2021, feng19_trainactive, roy18_deepactive, hekimoglu22_monocular, schmidt20_strategies, meyer19_radar, luo23_exploring} or a diversity-based approach \cite{liang22_diversity, luo23_exploring}, to determine the most informative training samples.  When selecting samples based on uncertainty, informativeness is measured by the predictive uncertainty, and the samples with the highest uncertainty are provided to the human annotators. When selecting samples based on diversity, scores are assigned based on spatial and temporal diversity \cite{liang22_diversity}. Both the uncertainty-based and diversity-based approaches have been used for bootstrapping 3D object detection, but the uncertainty-based approach has been shown to be more effective in practice. More recently, \cite{luo23_exploring} attempts to combine both strategies. Despite substantial work in active learning for image recognition and 2D object detection, exploration into active learning for 3D object detection is limited. ReBound helps support active learning by allowing users to filtering predicted annotations based on detection confidence score, which is useful for measuring uncertainty.

\section{ReBound: 3D Bounding-Box Re-Annotation}
In this section, we describe the functionality and software architecture of ReBound. Our code, documentation, and videos demonstrating our tool are available on \href{https://github.com/ajedgley/ReBound}{GitHub}.

\textbf{Data Format.} ReBound currently supports three different datasets: nuScenes, Waymo Open Dataset, and Argoverse 2.0. Annotations for these three datasets are converted into our generic ReBound format using a separate command line tool built in Python. Specifically, the generic data format captures the minimum information required to annotate a 3D bounding box (e.g. object center, box size, box rotation, and sensor extrinsics), which allows us to scale this format across multiple datasets. Users can support new datasets by extending our command line tool for their use case.

\textbf{Data Visualization.} The visualization tool has three windows: a control window, a point cloud viewer, and an RGB image viewer, as shown in Figure \ref{fig:rebound}. The control window allows the user to navigate through different frames in a driving log, switch between different camera views in the RGB image window, and filter through both ground truth annotations and model predictions. The point cloud viewer displays the point cloud corresponding to the current frame, as well as all ground truth annotations and predictions (if available) for that frame as wireframe cuboids. The user can rotate, translate, and zoom in/out to interact with the scene. We use Open3D as our rendering back-end as this natively supports 3D rendering on top of LiDAR sweeps and RGB images.

\textbf{Editing Tool.}
Users can directly click on a location in the point cloud window to add, edit, or delete an annotation. To edit a box’s properties, users must first click on a desired cuboid in the point cloud window. This will highlight the bounding box and display the position, rotation, size and annotation class of the selected box in the control window. These fields can all be directly updated, allowing the user to make precise changes. Users can also make coarse changes to the location of the selected box by clicking and dragging the box within the LiDAR viewer. To make it easier to interact with 3D objects, the mouse tool is restricted to two different control modes:

\begin{itemize}
    \item Horizontal Translation Mode: The user can articulate objects along the $X$ and $Y$ axes with the $Z$ axis locked
    \item Vertical Translation Mode: The user can click and drag objects along the $Z$ axis and rotate objects about the $Z$ axis. 
\end{itemize}

Finer-grained modifications can be manually entered in the control window. Users can also delete the selected bounding box, and create new bounding boxes with a single mouse-click. All bounding box transformations are updated in both the LiDAR and RGB image viewers in real-time. In practice, the editing tool can be used to update bounding boxes that are too large, are misaligned with the true object location (c.f. Figure \ref{fig:adjust}), update mis-classified objects, and add new categories like stop signs and traffic lights. 

\begin{figure}[t]
    \centering
    \includegraphics[scale=0.15]{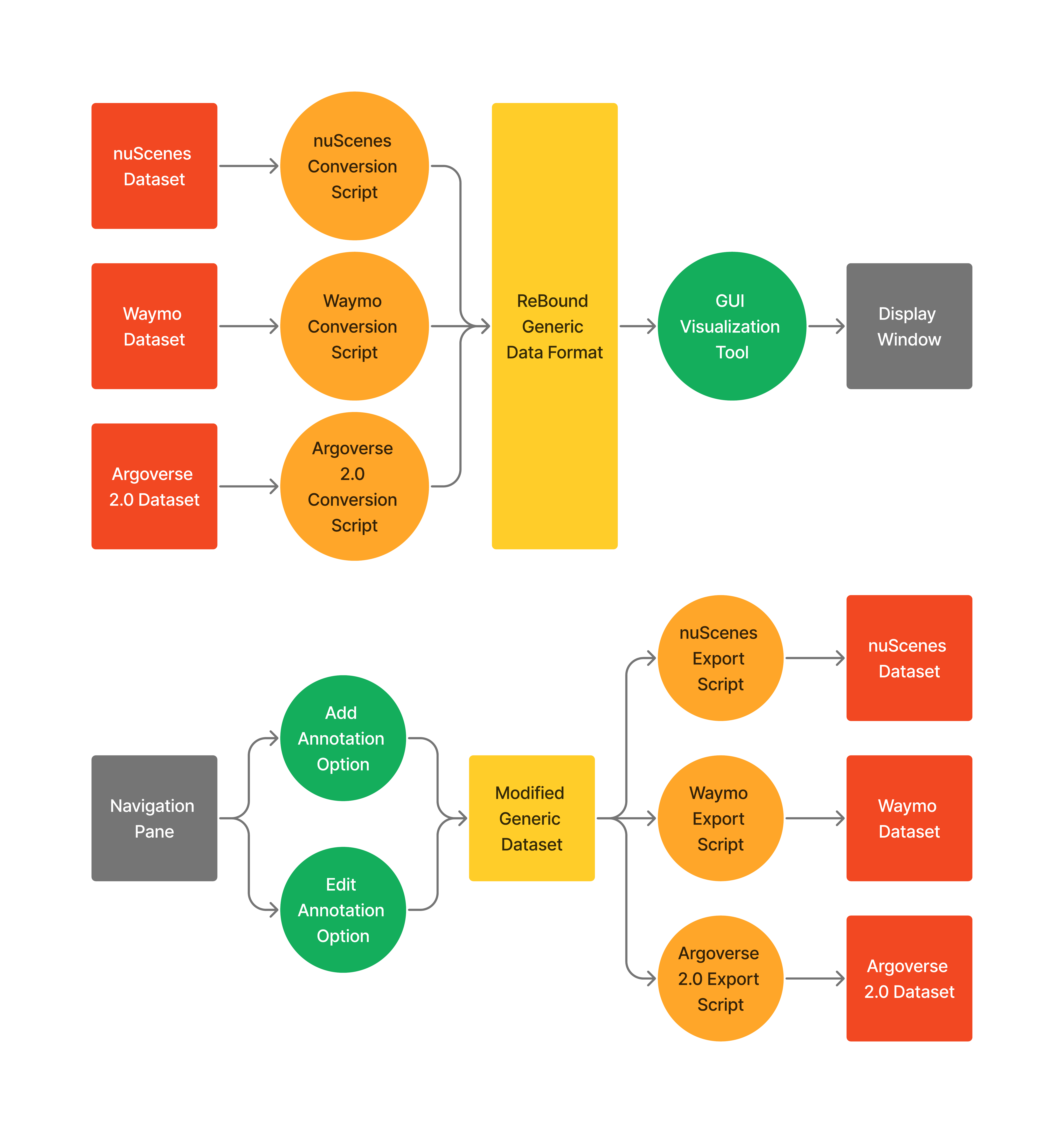}
    \caption{Our tool supports data conversion, visualization, and modification of the  nuScenes, Waymo and Argoverse 2.0 datasets. First, dataset-specific RGB images, LiDAR sweeps, sensor extrinsics, and annotations are converted to the ReBound data format using our conversion scripts and visualized in the GUI. Using this GUI, we can add, edit, or delete existing annotations. Lastly, we can export data from the ReBound data format back to the respective dataset-specific formats using the provided export scripts.}
\end{figure}

\newpage
\textbf{Exporting.}
Users can export updated annotations back to the original dataset-specific formats using a command line tool. Importantly, ReBound only exports the modified bounding box annotations back to the original format. Since ReBound does not modify the LiDAR sweeps or RGB images, we can directly use the data from the original dataset. We find that this dramatically increases data export speed. Interestingly, since we convert all datasets into a unified format, we can easily export annotations {\it between} datasets. Concretely, this means we can convert nuScenes annotations into the Argoverse 2.0 format using the ReBound format as an intermediary, making it easier to evaluate models across different datasets. Further, this can facilitate future research in model robustness and domain adaptation between different datasets. Similar to the data conversion tool that converts from dataset-specific formats to the ReBound format, users that wish to support a new dataset can extend our command line tool for their use case.

\section{Experiments}
In this section we present the results of our user survey to evaluate the effectiveness of our tool. 

\textbf{Ease of Use.}
We conducted surveys to evaluate the tool's ease of use in facilitating data conversion between data formats and bounding box adjustments among both people familiar and unfamiliar with autonomous vehicle datasets and 3D annotation tools.

We asked ten participants to perform 13 tasks (shown in Table \ref{tab:tasks}) over a video conference. Participants were shown a demonstration of the tool before being asked to complete a new, but related task using the tool. This method of demonstration allows us to avoid re-downloading the datasets and re-installing the software on a new computer for each trial. After completing all 13 tasks, users were asked to complete a questionnaire designed to rate different parts of the user experience on a scale of one to five (five is highest).

Based on our survey, we identified that the greatest challenge for users was translating and rotating bounding boxes. Both of these operations require fine-grained controls and navigating in a 3D space, which may be not be intuitive for some users. However these features also had high standard deviation compared to the other tasks, indicating that the utility of our tool may depend on prior use of 3D visualization and annotation tools. In general, participants stated that the application can be useful for autonomous vehicle research.

\begin{table}[b]

    \begin{adjustbox}{width=\columnwidth,center}
    \begin{tabular}{l|c|c}
    \toprule
    \mbox{Task List} & \mbox{Avg.} & \mbox{Std. Dev} \\ \hline
    \mbox{Convert nuScenes data to the ReBound generic type} & 4.1 & 0.74 \\
    \mbox{Enter annotation mode} & 4.2 & 0.79 \\
    \mbox{Add a new annotation bounding box} & 3.8 & 1.4 \\
    \mbox{Add a custom annotation type} & 4.7 & 0.67 \\
    \mbox{Select an existing bounding box} & 4.5 & 0.97 \\
    \mbox{Translate a bounding box} & 3.8 & 1.03 \\
    \mbox{Rotate a bounding box} & 3.6 & 1.17 \\
    \mbox{Change the label of a bounding box} & 4.8 & 0.42 \\
    \mbox{Delete a bounding box} & 4.8 & 0.63 \\
    \mbox{Use the control viewer to edit an annotation} & 4 & 1.15 \\
    \mbox{Save modified annotations} & 4.5 & 0.71 \\
    \mbox{Exit the application} & 4.5 & 0.53 \\
    \mbox{Export data back to the nuScenes data format} & 3.9 & 0.88 \\
    \bottomrule
    \end{tabular}
    \end{adjustbox}
    \caption{Our survey results show that users found it easier to create, delete, and modify annotations but found it more difficult to rotate, translate, and export bounding boxes from the ReBound data format to dataset-specific formats.}
    \label{tab:tasks}
\end{table}

\begin{figure}[t]
    \centering
    \includegraphics[scale=0.75]{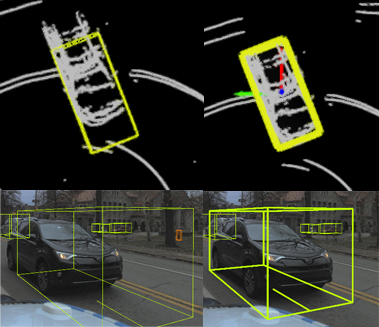}
    \caption{We visualize a ground truth annotation from the Argoverse 2.0 dataset (left). We find that the annotation is misaligned with the true car location. We adjust this ground truth annotation using ReBound (right). Visually inspecting the updated bounding box in both the LiDAR and RGB image viewer confirms that the updated annotation correctly localizes the car.}
    \label{fig:adjust}
\end{figure}

\textbf{Bounding Box Adjustments.} 
We visualize ground truth annotations in both the LiDAR and RGB image viewers. We are able to identify examples where the ground truth annotation is misaligned with the real object, as shown in Figure \ref{fig:adjust}, highlighting a potential application of our tool. This is practically meaningful as training 3D detectors with noisy labels can result in degraded model performance.

\section{Conclusion}
The academic community studies 3D object detection using publicly available AV datasets, which often provide LiDAR sweeps, RGB images, and 3D bounding box annotations. However, these datasets are limited by the annotations provided by their curators. In some cases, we find that these annotations may be incorrect or incomplete. Existing annotation tools cannot be easily used to update annotations for existing datasets, which makes it difficult to fix incorrect annotations, annotate new categories of interest, or iteratively improve 3D object detection models through active learning. In this paper, we propose ReBound, an open-source alternative that simplifies the process of bounding box re-annotation for existing datasets. We validate the utility of our tool through user surveys and find that users are able to rapidly add, modify, and delete 3D bounding box annotations.


\begin{acknowledgments}
Author James Purtilo was supported by N000142112821 while working on this project. This work is supported by the SEAM Lab at the University of Maryland and the Argo AI Center for Autonomous Vehicle Research at Carnegie Mellon University. 
\end{acknowledgments}

\newpage 
\bibliography{references}
\end{document}